\documentclass[runningheads]{llncs}

\pdfoutput=1
\usepackage[T1]{fontenc}
\def\doi#1{\href{https://doi.org/\detokenize{#1}}{\url{https://doi.org/\detokenize{#1}}}}
\usepackage{graphicx}
\usepackage{listings}
\usepackage[colorlinks=true,linkcolor=green,citecolor=blue]{hyperref}
\usepackage{algorithm}
\usepackage{algorithmic}
\usepackage{adjustbox}
\usepackage{array}
\usepackage{anyfontsize}
\usepackage{amssymb}  % http://ctan.org/pkg/amssymb
\usepackage{booktabs} 
\usepackage{changepage}
\usepackage{float}
\usepackage{gensymb}
\usepackage{graphicx}
\usepackage[all]{hypcap}
\usepackage{ltablex}
\usepackage{mathtools}
\usepackage{multicol}
\usepackage{multirow}
\usepackage{mwe}   % to get dummy images
\usepackage{pdflscape}
\usepackage{pifont}   % http://ctan.org/pkg/pifont
\usepackage{placeins}
\usepackage{rotating}
\usepackage{soul}
\usepackage{scalerel}
\usepackage{stackengine}
\usepackage{tabularx}
\usepackage{tabularray}
\usepackage{textcomp}
\usepackage{xcolor}

\def \name{SwIPE}

\usepackage[mathscr]{euscript}
\DeclareSymbolFont{rsfs}{U}{rsfs}{m}{n}
\DeclareSymbolFontAlphabet{\mathscrsfs}{rsfs}

\DeclarePairedDelimiter{\ceil}{\lceil}{\rceil}

\newcommand{\cmark}{\ding{51}}%

\newlength{\blob}
\newlength{\nameblob}
\begin{document}
%
% \title{Vector Prediction: a Surprisingly Robust Self-supervised 3D Pretraining Task for CT Segmentation}
\title{
SwIPE: Efficient and Robust Medical Image Segmentation with Implicit Patch Embeddings}
\titlerunning{\name}

% Non-anon Author List
\author{\vspace*{-0.5mm}
Yejia Zhang \and  
Pengfei Gu \and   
Nishchal Sapkota \and  
Danny Z. Chen     
}
%index{Zhang, Yejia}
%index{Gu, Pengfei}
%index{Sapkota, Nishchal}
%index{Chen, Danny}

\institute{\vspace*{-0.25mm}
University of Notre Dame, Notre Dame IN 46556, USA \\
\email{\{yzhang46,pgu,nsapkota,dchen\}@nd.edu}}
\authorrunning{Zhang et al.}

\maketitle            
\begin{abstract}  % 150 - 250 words
\vspace{-3.4mm}
Modern medical image segmentation methods primarily use \textit{discrete} representations in the form of rasterized masks to learn features and generate predictions. 
Although effective, this paradigm is spatially inflexible, scales poorly to higher-resolution images, and lacks direct understanding of object shapes. 
% To address these limitations, some recent works utilized implicit neural representations (INRs) to learn \textit{continuous} representations for segmentation, but often directly adopted components designed for 3D shape reconstruction. 
% These formulations also constrained themselves to either point-based or global contexts which lack contextual understanding or local fine-grained details, respectively --- both essential for accurate segmentation.
To address these limitations, some recent works utilized implicit neural representations (INRs) to learn \textit{continuous} representations for segmentation.
However, these methods often directly adopted components designed for 3D shape reconstruction. 
More importantly, these formulations were also constrained to either point-based or global contexts, lacking contextual understanding or local fine-grained details, respectively---both critical for accurate segmentation.
To remedy this, we propose a novel approach, \textbf{SwIPE} (\underline{S}egmentation \underline{w}ith \underline{I}mplicit \underline{P}atch \underline{E}mbeddings), that leverages the advantages of INRs and predicts shapes at the patch level---rather than at the point level or image level---to enable both accurate local boundary delineation and global shape coherence.
% by modeling shapes at an image patch level.
% maintaining fine-grained local information and global shape coherence.
Extensive evaluations on two tasks (2D polyp segmentation and 3D abdominal organ segmentation) show that SwIPE significantly improves over recent implicit approaches and outperforms state-of-the-art discrete methods with over 10x fewer parameters.
Our method also demonstrates superior data efficiency and improved robustness to data shifts across image resolutions and datasets.
Code is available on \href{https://github.com/charzharr/miccai23-swipe-implicit-segmentation}{Github}.

\vspace{-2.5mm}
\end{abstract}

%%%%%%% ------------------------------------------------------------------------- %%%%%%%
%%%%%%% ------------------------------------------------------------------------- %%%%%%%

\section{Introduction}  \label{sec:1}
\vspace{-1.5mm}

Segmentation is a critical task in medical image analysis.
%Contemporary 
Known approaches mainly utilize \textit{discrete} data representations (e.g., rasterized label masks) with convolutional neural networks (CNNs) \cite{isensee2021nnu,fan2020pranet,ronneberger2015unet,gu2021kcbac} or Transformers \cite{hatamizadeh2022unetr,hassani2021CCT} to classify image entities in a bottom-up manner.
While undeniably effective, this paradigm suffers from \textit{two primary limitations}.
(1) These approaches have limited spatial flexibility and poor computational scaling.
Retrieving predictions at higher resolutions would require either increasing the input size, which decreases performance and incurs quadratic or cubic memory increases, or interpolating output predictions, which introduces discretization artifacts.
(2) Per-pixel or voxel learning inadequately models object shapes/boundaries, which are central to both robust computer vision methods and our own visual cortical pathways~\cite{pasupathy2015neuralprimatebasis}. 
This often results in predictions with unrealistic object shapes and locations~\cite{raju2022dissm}, especially in settings with limited annotations and out-of-distribution data.

Instead of segmenting structures with \textit{discrete} grids, we explore the use of Implicit Neural Representations (INRs) which employ \textit{continuous} representations to compactly capture coordinate-based signals (e.g., objects in images). 
INRs represent object shapes with a parameterized function $f_\theta: (\textbf{p}, \textbf{z}) \rightarrow [0, 1]$ that maps continuous spatial coordinates $\textbf{p} = (x, y, z)$, $x, y, z \in [-1,1]$ and a shape embedding vector $\textbf{z}$ to occupancy scores.
This formulation enables direct modeling of object contours as the decision boundary of $f_\theta$, superior memory efficiency~\cite{dupont2021coincompressionINR}, and smooth predictions at arbitrary resolutions that are invariant to input size. 
INRs have been adopted in the vision community for shape reconstruction~\cite{Chibane2020IFNet,Park2019DeepSDF,Mescheder2018OccNet,chabra2020deeplocalshape}, texture synthesis~\cite{oechsle2019texture}, novel view synthesis~\cite{Mildenhall2020NeRFRS}, and segmentation~\cite{hu2022ifanet}.
Medical imaging studies have also used INRs to learn organ templates \cite{yang2022implicitatlas}, synthesize cell shapes \cite{wiesner2022inrCellShape}, and reconstruct radiology images \cite{shen2022nerp}.

The adoption of INRs for medical image segmentation, however, has been limited where most existing approaches directly apply pipelines designed for 3D reconstruction to images.
These works emphasize either global embeddings $\textbf{z}$ or point-wise ones.
OSSNet \cite{Reich2021OSSNetME} encodes a global embedding from an entire volume and an auxiliary local image patch to guide voxel-wise occupancy prediction. 
Although global shape embeddings facilitate overall shape coherence, they neglect the fine-grained details needed to delineate local boundaries.
The local patches partially address this issue but lack contextual understanding beyond the patches and neglect mid-scale information. 
% Small image patch encodings are also incorporated since global embeddings cannot accurately model local details (e.g., boundaries).
% Using cropped patches, however, fails to leverage neighboring context.
% However, inputting small image patches fails to leverage neighboring context and may not adequately encode mid-scale information. 
In an effort to enhance local acuity and contextual modeling, IFA~\cite{hu2022ifanet}, IOSNet~\cite{Khan2022IOSNet}, and NUDF~\cite{Srensen2022NUDF} each extract a separate embedding for every input coordinate by concatenating point-wise features from multi-scale CNN feature maps.
Although more expressive, point-wise features still lack sufficient global contextual understanding and suffer from the same unconstrained prediction issues observed in discrete segmentation methods.
Moreover, these methods use components designed for shape reconstruction---a domain where synthetic data is abundant and the modeling of texture, multi-class discrimination, and multi-scale contexts are less crucial.
% Further, DISSM~\cite{raju2022dissm} takes a separate approach and uses INRs in a Statistical Shape Model schema to construct templates to be non-rigidly aligned and refined. 
% Although effective, this approach is not end-to-end and contains components such as non-rigid alignment and boundary refinement that are difficult to train and operate expensively in discrete space.

To address these limitations, we propose \textbf{SwIPE (\underline{S}egmentation \underline{w}ith \underline{I}mplicit \underline{P}atch \underline{E}mbeddings)} which learns continuous representations of foreground shapes at the patch level. 
% Operating on patches rather than the entire image or points better enables both local boundary details and global shape coherence.
By decomposing objects into parts (i.e., patches), we aim to enable both accurate local boundary delineation and global shape coherence. 
This also improves model generalizability and training efficiency since local curvatures often reoccur across classes or images. 
%(see Table~\autoref{tab:2}). 
% which decomposes images into a uniform grid of patches and objects into parts. 
% Our method decomposes images into a uniform grid of patches and objects into parts, enabling more accurate local boundary delineation and regional shape coherence.
% This also improves model generalization since 
% and avoid polarization of patch embeddings toward either local or global features
SwIPE first encodes an image into descriptive patch embeddings and then decodes the point-wise occupancies using these embeddings.
To avoid polarization of patch embeddings toward either local or global features in the encoding step, we introduce a context aggregation mechanism that fuses multi-scale feature maps and propose a \textbf{Multi-stage Embedding Attention (MEA)} module to dynamically extract relevant features from all scales. 
This is driven by the insight that different object parts necessitate variable focus on either global/abstract (important for object interiors) or local/fine-grained information (essential around object boundaries). 
To enhance global shape coherence across patches in the decoding step, we augment local embeddings with global information and propose \textbf{Stochastic Patch Overreach (SPO)} to improve continuity around patch boundaries.
Comprehensive evaluations are conducted on two tasks (2D polyp and 3D abdominal organ segmentation) across four datasets.
SwIPE outperforms the best-known implicit methods (+6.7\% \& +4.5\% F1 on polyp and abdominal, resp.) and beats task-specific discrete approaches (+2.5\% F1 on polyp) with 10x fewer parameters.
We also demonstrate SwIPE's superior model \& data efficiency in terms of network size \& annotation budgets, and greater robustness to data shifts across image resolutions and datasets. Our main \textbf{contributions} are as follows.
\vspace*{-2.4mm}
\begin{enumerate}
    \item Away
    %Deviating 
    from discrete representations, we are the first to showcase the merits of patch-based implicit neural representations for medical image segmentation. 
    % Our method, SwIPE (\underline{S}egmentation \underline{w}ith \underline{I}mplicit \underline{P}atch \underline{E}mbeddings), can output predictions of arbitrary resolutions with favorable memory scaling and directly model object shapes.
    % , and makes significant design improvements over previous implicit-based segmentation methods.
    \item We propose a new efficient attention mechanism, Multi-stage Embedding Attention (MEA), to improve contextual understanding during the encoding step, and Stochastic Patch Overreach (SPO) to address boundary continuities during occupancy decoding.
    % demonstrate the importance of contextual understanding for modeling local shapes and introduce an efficient attention mechanism, multi-stage embedding attention (MEA), that dynamically extracts information from multiple feature abstraction levels.
    % For accurate decoding, we propose Stochastic Patch Overreach (SPO) that enhances boundary continuity and cross-patch understanding.
    \item We perform detailed evaluations of SwIPE and its components on two tasks (2D polyp segmentation and 3D abdominal organ segmentation). We not only outperform 
    %recent 
    state-of-the-art implicit and discrete methods, but also 
    %demonstrate 
    yield improved data \& model efficiency and better robustness to data shifts.
\end{enumerate}

% To obtain descriptive patch embeddings, we introduce a context aggregation module to fuse multi-scale feature maps and propose an efficient attention mechanism to dynamically weigh multi-scale features.
% This is driven by the insight that different parts of an object necessitate different degrees of attention on global/abstract and local/fine-grained details.
% Further, to augment global shape coherence across patches, we adopt three mechanisms to improve continuity at patch boundaries and regularize the global shape.
% For robust patch prediction with adequate global shape regularization, we propose three mechansisms to help achieve this.
% We introduce a stochastic patch overreach scheme to alleviate boundary discontinuity between patches and faciliate contextual understanding beyond a local scope. 
% We also condition local patches on global embeddings. 

\vspace{-5.3mm}
\section{Methodology} \label{sec:2}
\vspace{-2.1mm}

\begin{figure}[!t]
    \centering
    % \begin{minipage}[b]{0.8\linewidth}
      \centering
      % \centerline{\includegraphics[width=\linewidth]{assets/Asset 15.pdf}}
      \includegraphics[width=1.0\linewidth]{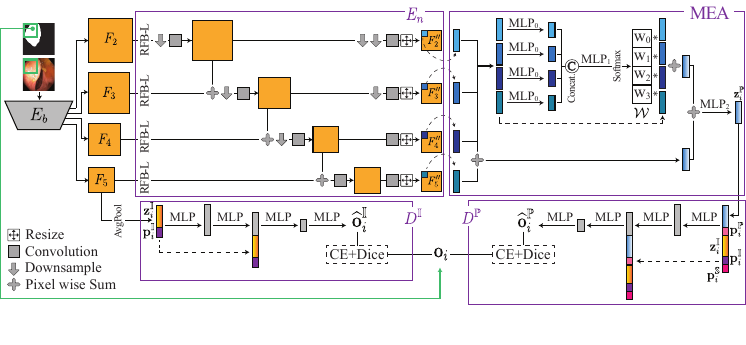}
    %  \vspace{2.0cm}
    %  \centerline{}\medskip
    % \end{minipage}
    %
    \vspace*{-14mm}
    \caption{\label{fig:1}
    At a high level, SwIPE first encodes an input image into patch $\textbf{z}^\mathbb{P}$ and image $\textbf{z}^\mathbb{I}$ shape embeddings, and then employs these embeddings along with coordinate information $\textbf{p}$ to predict class occupancy scores via the patch $\textbf{D}^\mathbb{P}$ and image $\textbf{D}^\mathbb{I}$ decoders.
    }
    \vspace*{-5mm}
    % SwIPE framework overview: an input image is initially encoded into patch $\textbf{z}^\mathbb{P}$ and image $\textbf{z}^\mathbb{I}$ shape embeddings. These embeddings, coupled with coordinate information $\textbf{p}$, are then utilized by the patch $\textbf{D}^\mathbb{P}$ and image $\textbf{D}^\mathbb{I}$ decoders to predict point-by-point class occupancy scores.
        
        % At a high level, SwIPE first encodes an input image into patch $\textbf{z}^\mathbb{P}$ and image $\textbf{z}^\mathbb{I}$ shape embeddings. 
        % The patch decoder $\textbf{D}^\mathbb{P}$ and image decoder $\textbf{D}^\mathbb{I}$ then use these embeddings along with coordinate information $\textbf{p}$ to predict class occupancy scores point by point.
    
        % An overview of how SwIPE encodes an input image into patch $\textbf{z}^\mathbb{P}$ and image $\textbf{z}^\mathbb{I}$ shape embeddings, which are then decoded at each coordinate $\textbf{p}$ to class occupancy scores with the patch decoder $\textbf{D}^\mathbb{P}$ and image decoder $\textbf{D}^\mathbb{I}$.
        
        % . The encoder components (top of figure) map an input image to patch $\textbf{z}^\mathbb{P}$ and image $\textbf{z}^\mathbb{I}$ shape embeddings. The patch decoder $\textbf{D}^\mathbb{P}$ and global image decoder $\textbf{D}^\mathbb{I}$ are conditioned on these embeddings along with relevant coordinates $\textbf{p}$ to predict class occupancy scores at each coordinate.
    % \vspace*{-5mm}
    
\end{figure}

The core idea of SwIPE (overviewed in Fig.~\ref{fig:1}) is to use patch-wise 
%implicit neural representations (INRs) 
INRs for semantic segmentation.
To formulate this, we first discuss the shift from discrete to implicit segmentation, then delineate the intermediate representations needed for such segmentation, and overview the major components involved in obtaining these representations.
Note that for the remainder of the paper, we present formulations for 2D data but the descriptions are conceptually congruous in 3D.

In a typical discrete segmentation setting with $C$ classes, an input image $\textbf{X}$ is mapped to class probabilities with the same resolution $f: \textbf{X} \in \mathbb{R}^{H \times W \times 3} \rightarrow \hat{\textbf{Y}} \in \mathbb{R}^{H \times W \times C}$. 
Segmentation with INRs, on the other hand, maps an image $\textbf{X}$ and a continuous image coordinate $\textbf{p}_i = (x, y)$, $x, y \in [-1,1]$, to the coordinate's class-wise occupancy probability $\hat{\textbf{o}}_i \in \mathbb{R}^C$, yielding $f_\theta: (\textbf{p}_i, \textbf{X}) \rightarrow \hat{\textbf{o}}_i$, where $f_\theta$ is parameterized by a neural network with weights $\theta$.
As a result, predictions of arbitrary resolutions can be obtained by modulating the spatial granularity of the input coordinates.
This formulation also enables the direct use of discrete pixel-wise losses like Cross Entropy or Dice with the added benefit of boundary modeling.
Object boundaries are represented as the zero-isosurface in $f_\theta$'s prediction space or, more elegantly, $f_\theta$'s decision boundary.
% As a result, from a fixed size $X$, predictions of arbitrary resolutions can be obtained by modulating the spacial granularity of input coordinates.
% This formulation also allows for seamless integration with existing discrete pipelines with pixel-wise losses like Cross Entropy or Dice, but with the added benefit of robust boundary modeling in the form of the zero-isosurface from $D$'s prediction space.
% Thus, on a high level, our method operates on point and class-occupancy pairs via $f_\theta$ rather than pixel-wise correspondences between discrete images and masks.

SwIPE builds on the INR segmentation setting (e.g., in~\cite{Khan2022IOSNet}), but operates on patches rather than on points or global embeddings (see Tab.~\ref{tab:1} \& left of Tab.~\ref{tab:3} for empirical justifications) to better enable both local boundary details and global shape coherence.
This involves two main steps: (1) encode shape embeddings from an image, and (2) decode occupancies for each point while conditioning on its corresponding embedding(s).
In our case, $f_\theta$ includes an encoder $E_b$ (or backbone) that extracts multi-scale feature maps from an input image, a context aggregation module $E_n$ (or neck) that aggregates the feature maps into vector embeddings for each patch, and MLP decoders $D^\mathbb{P}$ (decoder for local patches where $\mathbb{P}$ is for patch) \& $D^\mathbb{I}$ (decoder for entire images where $\mathbb{I}$ is for image) that output smoothly-varying occupancy predictions given embedding \& coordinate pairs.
To encode patch embeddings in step (1), $E_b$ and $E_n$ map an input image $\textbf{X}$ to a global image embedding $\textbf{z}^\mathbb{I}$ and a matrix $\textbf{Z}^\mathbb{P}$ containing a local patch embedding $\textbf{z}^\mathbb{P}$ at each planar position. 
For occupancy decoding in step (2), $D^\mathbb{P}$ decodes the patch-wise class occupancies $\textbf{o}_i^\mathbb{P}$ using relevant local and global inputs while $D^\mathbb{I}$ predicts occupancies $\textbf{o}_i^\mathbb{I}$ for the entire image using only image coordinates $\textbf{p}_i^\mathbb{I}$ and the image embedding $\textbf{z}^\mathbb{I}$.
% For occupancy decoding in (2), $D^\mathbb{P}$ decodes a patch coordinate $\textbf{p}_i^\mathbb{P}$ and its corresponding patch embedding $\textbf{z}_i^\mathbb{P}$ to patch class occupancies $D^\mathbb{P}: \textbf{p}_i^\mathbb{P}, \textbf{z}_i^\mathbb{P} \rightarrow \hat{\textbf{o}}_i^\mathbb{P}$. 
% Similarly, image occupancy decoding can be written as $D^\mathbb{I}: \textbf{p}_i^\mathbb{I}, \textbf{z}_i^\mathbb{I} \rightarrow \hat{\textbf{o}}_i^\mathbb{I}$ and serves as an auxiliary task to learn global shapes across patches.
Below, we detail the encoding of image \& patch embeddings (Sec.~\autoref{sec:2-1}), the point-wise decoding process (Sec.~\autoref{sec:2-2}), and the training procedure for SwIPE (Sec.~\autoref{sec:2-3}).

% Since our method operates on patch embeddigns
% $f_\theta$ in SwIPE consists of an encoder $E_b$ (or backbone) that extracts multi-scale feature maps from an input image, a context aggregation module $E_n$ (or neck) that aggregates the feature maps into vector embeddings for each patch, and a decoder $D$ that outputs smoothly-varying occupancy predictions for each patch.

% Below, we detail the components used to extract patch embeddings from image $X$ (see~\autoref{sec:2-1}), the mechanisms for decoding patch shapes from patch embeddings (see~\autoref{sec:2-2}), and the training procedure for SwIPE (see~\autoref{sec:2-3}). 

\vspace{-4mm}
\subsection{Image Encoding and Patch Embeddings} \label{sec:2-1}
\vspace{-1.5mm}

The encoding process utilizes the backbone $E_b$ and neck $E_n$ to obtain a global image embedding $\textbf{z}^\mathbb{I}$ and a matrix $\textbf{Z}^\mathbb{P}$ of patch embeddings.
% $E_n \circ E_b: \textbf{X} \rightarrow \textbf{z}^\mathbb{I}, \textbf{Z}^\mathbb{P}$.
We define an image patch as an isotropic grid cell (i.e., a square in 2D or a cube with identical spacing in 3D) of length $S$ from non-overlapping grid cells over an image.
Thus, an image $\textbf{X} \in \mathbb{R}^{H \times W \times 3}$ with a patch size $S$ will produce $\ceil[\big]{\frac{H}{S}} \cdot \ceil[\big]{\frac{W}{S}}$ patches.
For simplicity, we assume that the image dimensions are evenly divisible by $S$.
% written as $E_n \circ E_b: \textbf{X} \rightarrow \textbf{z}^\mathbb{I}, \textbf{Z}^\mathbb{P}$. where $d$ is the embedding dimension, $\textbf{z}^\mathbb{I} \in \mathbb{R}^d$, and $\textbf{Z}^\mathbb{P} \in \mathbb{R}^{\frac{H}{S} \times \frac{W}{S} \times d}$.
% $\textbf{z}_{rc} \in \textbf{Z}$ indexes the patch embedding at the $r$th row and $c$th column.

A fully convolutional \textbf{encoder backbone} $E_b$ (e.g., Res2Net-50~\cite{gao2019res2net}) is employed to generate multi-scale features from image $\textbf{X}$.
The entire image is processed as opposed to individual crops \cite{jiang2020localimplicitgrid,chabra2020deeplocalshape,Reich2021OSSNetME} to leverage larger receptive fields and integrate intra-patch information.
Transformers \cite{hassani2021CCT} also model cross-patch relations and naturally operate on patch embeddings, but are data-hungry and lack helpful spatial inductive biases (we affirm this in Sec.~\autoref{sec:3-5}).
$E_b$ outputs four multi-scale feature maps from the last four stages, $\{\textbf{F}_n\}_{n=2}^{5}$ ($\textbf{F}_n \in \mathbb{R}^{C_n \times H_n \times W_n}$, $H_n = \frac{H}{2^n}$, $W_n = \frac{W}{2^n}$).
% To obtain $\textbf{Z}^\mathbb{P}$, we prioritize data efficiency and contextual understanding in our encoder design for accurate segmentation with limited medical data.
% For $E_b$ and $E_n$, we elect to use fully convolutional encoder components that operate on the entire input image to encode patch embeddings.
% We avoid directly inputting crops like other patch-based 3D reconstruction approaches~\cite{jiang2020localimplicitgrid,chabra2020deeplocalshape} since they preclude helpful context beyond patches.
% For better, context modeling, Transformers \cite{hassani2021CCT} were also considered since they innately represent image patches as vector embeddings and are effective in modeling long-range relations, but self-attention requires abundant data and lack helpful spacial inductive biases (see backbone studies in Section~\autoref{sec:3-5}).
% Thus, we employ a multi-stage convolutional \textbf{encoder backbone} $E_b$ to generate multiple levels of features $\{\textbf{F}_n\}_{n=2}^{5}$ from image $\textbf{X}$ where $\textbf{F}_n \in \mathbb{R}^{C_n \times H_n \times W_n}$, $H_n = \frac{H_n}{2^n}$.

The \textbf{encoder neck} $E_n$ aggregates $E_b$'s multi-scale outputs $\{\textbf{F}_n\}_{n=2}^{5}$ to produce $\textbf{z}^\mathbb{I}$ (the shape embedding for the entire image) and $\textbf{Z}^\mathbb{P}$ (the grid of shape embeddings for patches).
The feature maps are initially fed into a modified Receptive Field Block \cite{liu2018receptivefieldblockRFB} (dubbed RFB-L or RFB-Lite) that replaces symmetric convolutions with a series of efficient asymmetric convolutions 
(e.g., $(3\times3)$ $\rightarrow$ $(3\times1) + (1\times3)$).
% Like modern object detection schemas \cite{zhu2021tphyolov5}, 
The context-enriched feature maps are then fed through multiple cascaded aggregation and downsampling operations (see $E_n$ in Fig.~\ref{fig:1}) to obtain four multi-stage intermediate embeddings with identical shapes, $\{\textbf{F}_n'\}_{n=2}^{5} \in \mathbb{R} ^ {\frac{H}{32} \times \frac{W}{32} \times d}$.
% Similar to modern object detection schemas \cite{zhu2021tphyolov5}, we adopt an \textbf{encoder neck} $E_n$ to aggregate multi-scale features and improve contextual understanding (see Figure~\ref{fig:1}).
% All multiscale feature maps $\{\textbf{F}_n\}_{n=2}^{5}$ from $E_b$ are initially fed into a modified Receptive Field Block \cite{liu2018receptivefieldblockRFB} (dubbed RFB-L or RFB-Lite) that replaces symmetric convolutions with a series of efficient asymmetric convolutions 
% (e.g., $(3\times3)$ $\rightarrow$ $(3\times1) + (1\times3)$).
% % (e.g., $(3\times3)$ $\rightarrow$ $(3\times1) + (1\times3)$, or $(3\times3\times3)$ $\rightarrow$ $(3\times1\times1) + (1\times3\times1) + (1\times1\times3)$).
% The resulting augmented feature maps are then fed through multiple cascaded aggregation and downsampling operations to obtain four multi-stage feature maps with identical shapes $\{\textbf{F}_n'\}_{n=2}^{5} \in \mathbb{R} ^ {\frac{H}{32} \times \frac{W}{32} \times d}$.
% The vector at each planar position of $\textbf{F}_n'$ represents an intermediate embedding for a shape centered at that feature map position. 

To convert the intermediate embeddings $\{\textbf{F}_n'\}_{n=2}^{5}$ to patch embeddings $\textbf{Z}^\mathbb{P}$, we first resize them to $\textbf{Z}^\mathbb{P}$'s final shape via linear interpolation to produce $\{\textbf{F}_n''\}_{n=2}^{5}$, which contain low-level ($\textbf{F}_2''$) to high-level ($\textbf{F}_5''$) information.
Resizing enables flexibility in designing appropriate patch coverage, which may differ across tasks due to varying structure sizes and shape complexities.
Note that this is different from the interpolative sampling in~\cite{Khan2022IOSNet} and more similar to~\cite{hu2022ifanet}, except the embeddings' spatial coverage in SwIPE are larger and adjustable.
To prevent the polarization of embeddings toward either local or global scopes, we propose a \textbf{multi-stage embedding attention} (MEA) module to enhance representational power and enable dynamic focus on the most relevant abstraction level for each patch. 
Given four intermediate embedding vectors $\{\textbf{e}_n\}_{n=2}^{5}$ from corresponding positions in $\{\textbf{F}_n''\}_{n=2}^{5}$, we compute the attention weights via $\mathcal{W} = Softmax(MLP_1(cat(MLP_0(\textbf{e}_2), MLP_0(\textbf{e}_3), MLP_0(\textbf{e}_4), MLP_0(\textbf{e}_5))))$, where $\mathcal{W} \in \mathbb{R}^{4}$ is a weight vector, $cat$ indicates concatenation, and $MLP_0$ is followed by a ReLU activation. 
The final patch embedding is obtained by $\textbf{z}^\mathbb{P} = MLP_{2}(\sum^{5}_{n=2}  \textbf{e}_n + \sum^{5}_{n=2} w_{n-2} \cdot \textbf{e}_n)$, where $w_i$ is the $i$th weight of $\mathcal{W}$.
Compared to other spatial attention mechanisms like CBAM~\cite{woo2018cbam}, our module separately aggregates features at each position across multiple inputs and predicts a proper probability distribution in $\mathcal{W}$ instead of an unconstrained score. 
The output patch embedding matrix $\textbf{Z}^\mathbb{P}$ is populated with $\textbf{z}^\mathbb{P}$ at each position and models shape information centered at the corresponding patch in the input image (e.g., if $S=32$, $\textbf{Z}^\mathbb{P}[0,0]$ encodes shape information of the top left patch of size $32\times32$ in $\textbf{X}$).
Finally, $\textbf{z}^\mathbb{I}$ is obtained by average-pooling $\textbf{F}'_5$ into a vector.
% $\textbf{e} = MLP_2(\sum_{n=2}^{5} \textbf{e_n} + \mathcal{W}T )$.
% Note that this is similar in spirit to spatial attention in CBAM~\cite{woo2018cbam}, but with probability distribution $\mathcal{W}$ modeling multi-stage relevancy and involving multiple feature sources.

% For $E_d$ and $E_n$, we elect to use fully convolutional encoder components that operate on the entire input image to encode patch embeddings.
% However, there are other approaches to obtaining a descriptive embedding $\textbf{z}$. 
% Recent patch-based methods for 3D reconstruction \cite{jiang2020localimplicitgrid,chabra2020deeplocalshape} directly feed in local patches after cropping to extract embeddings.
% However, this faces similar limitations as the local encodings method in OSSNet \cite{Reich2021OSSNetME} by failing to leverage helpful context beyond the patch. 
% To improve context across patches, another natural approach is to adopt Transformers which already model image patches as vector embeddings and are effective in modeling long-range relations.
% Given limited medical image data, however, the self-attention mechanism lacks helpful inductive biases such as local bias, equivariant processing, and weight-sharing (we empirically affirm this in our backbone studies, see~\autoref{sec:3-5}). 
% Thus, we find a convolutional backbone to be the most suitable given their data, parameter, and computational efficiency.

\vspace{-3mm}
\subsection{Implicit Patch Decoding} \label{sec:2-2}
\vspace{-1.5mm}

% To predict patch-wise occupancies with decoder $D^\mathbb{P}$ ($\mathbb{P}$ for patch), 
Given an image coordinate $\textbf{p}^\mathbb{I}_i$ and its corresponding patch embedding $\textbf{z}_i^\mathbb{P}$, the patch-wise occupancy can be decoded with decoder $D^\mathbb{P}: (\textbf{p}^\mathbb{P}_i, \textbf{z}_i^\mathbb{P}) \rightarrow \hat{\textbf{o}}_i^\mathbb{P}$, where $D^\mathbb{P}$ is a small MLP and 
$\textbf{p}^\mathbb{P}_i$ is the patch coordinate with respect to the patch center $\textbf{c}_i$ associated with $\textbf{z}_i^\mathbb{P}$ and is obtained by $\textbf{p}^\mathbb{P}_i = \textbf{p}^\mathbb{I}_i - \textbf{c}_i$.
But, this design leads to poor global shape predictions and discontinuities around patch borders.
% Despite good local precision, decoding each patch independently in this manner leads to poor global shape coherence and discontinuities around patch borders.

To encourage better \textbf{global shape coherence}, we also incorporate a global image-level decoder $D^\mathbb{I}$.
This image decoder, $D^\mathbb{I}: (\textbf{p}^\mathbb{I}_i, \textbf{z}^\mathbb{I}) \rightarrow \hat{\textbf{o}}_i^\mathbb{I}$, predicts occupancies for the entire input image.
To distill higher-level shape information into patch-based predictions, we also condition $D^\mathbb{P}$'s predictions on $\textbf{p}_i^\mathbb{I}$ and $\textbf{z}^\mathbb{I}$. 
Furthermore, we find that providing the \textbf{source coordinate} gives additional spatial context for making location-coherent predictions.
In a typical segmentation pipeline, the input image $\textbf{X}$ is a resized crop from a source image and we find that giving the coordinate $\textbf{p}_i^\mathbb{S}$ ($\mathbb{S}$ for source) from the original uncropped image improves performance on 3D tasks since the additional positional information may be useful for predicting recurring structures.
Our enhanced formulation for patch decoding can be described as $D^\mathbb{P}: (\textbf{p}^\mathbb{P}_i, \textbf{z}^\mathbb{P}_i, \textbf{p}^\mathbb{I}_i, \textbf{z}^\mathbb{I}, \textbf{p}^\mathbb{S}_i) \rightarrow \hat{\textbf{o}}_i^\mathbb{P}$. 
% To encourage better \textbf{global shape coherence}, we also incorporate a global image-level decoder $D^\mathbb{I}$.
% The image decoder, $D^\mathbb{I}: \textbf{p}^\mathbb{I}_i, \textbf{z}^\mathbb{I} \rightarrow \hat{\textbf{o}}_i$, predicts occupancies for the entire input image given an image coordinate $\textbf{p}^\mathbb{I}_i$ and a global embedding $\textbf{z}^\mathbb{I}$ where $\textbf{z}^I$ is obtained by average-pooling $\textbf{F}'_5$ into a vector.
% To distill higher-level shape information into patch-based prediction, we also condition $D^\mathbb{P}$ predictions on $\textbf{p}_i^\mathbb{I}$ and $\textbf{z}^\mathbb{I}$. 
% Furthermore, we find that providing the \textbf{source coordinate} gives additional spacial context for making location-coherent predictions.
% In usual data pipelines, the input image $\textbf{X}$ is a resized crop from a source image and giving the coordinate $p_i^\mathbb{S}$ ($\mathbb{S}$ for source) from the original image particularly helps for 3D tasks.
% Finally, we have $D^\mathbb{P}: \textbf{p}^\mathbb{P}_i, \textbf{z}^\mathbb{P}, \textbf{p}^\mathbb{I}_i, \textbf{z}^\mathbb{I}, \textbf{p}^\mathbb{S} \rightarrow \hat{\textbf{o}}_i$. 

To address discontinuities at patch boundaries, we propose a training technique called \textbf{Stochastic Patch Overreach} (SPO) which forces patch embeddings to make predictions for coordinates in neighboring patches.
For each patch point and embedding pair ($\textbf{p}^\mathbb{P}_i, \textbf{z}^\mathbb{P}_i$), we create a new pair ($\textbf{p}^\mathbb{P}_i{}', \textbf{z}^\mathbb{P}_i{}'$) by randomly selecting a neighboring patch embedding and updating the local point to be relative to the new patch center.
This regularization is modulated by the set of valid choices to select a neighboring patch (\textit{connectivity}, or \textit{con}) and the number of perturbed points to sample per batch point (\textit{occurrence}, or $N_{\text{SPO}}$).
$con$=$4$ means all adjoining patches are neighbors while $con$=$8$ includes corner patches as well. 
Note that SPO differs from the regularization in \cite{chabra2020deeplocalshape} since no construction of a KD-Tree is required and we introduce a tunable stochastic component which further helps with regularization under limited-data settings.

% \subsubsection{Improving Boundaries}

% The duality of boundary decoding: 1) unsmooth at patch boundaries, 2) too smooth near object boundaries within the patch. 

% To achieve robust patch prediction with adequate global shape regularization, we propose three mechanisms. Firstly, we introduce a stochastic patch overreach scheme to alleviate boundary discontinuity between patches and facilitate contextual understanding beyond a local scope. Secondly, we condition local patches on global embeddings. Lastly, we apply multi-scale self-supervision to improve local and global shape consistency.

\begin{table*}[!t]
\begin{center}
\caption{\label{tab:1}
\textbf{Overall results versus the state-of-the-art}.
Starred* items indicate a state-of-the-art discrete method for each task.
The Dice columns report foreground-averaged scores and standard deviations ($\pm$) across 6 runs (6 different seeds were used while train/val/test splits were kept consistent).
\vspace*{-3mm}
}

\resizebox{0.8\textwidth}{!}{
\begin{tblr}{
    columns={colsep=3pt},
    colspec={l c c c  || l c c c },
    row{1} = {gray!50!black!15},
    row{3,7} = {gray!20!black!5}
    }
\hline 
\SetCell[c=4]{c} 2D Polyp Sessile & & & & \SetCell[c=4]{c} 3D CT BCV \\
\cline{1-4} \cline{5-8}
Method & Params (M) & FLOPs (G) & Dice (\%) & Method & Params (M) & FLOPs (G) & Dice (\%) \\
\hline 
\SetCell[c=8]{l} \textit{Discrete Approaches} \\
\hline 
\makebox[\nameblob][l]{U-Net$_{15}$}
\makebox[\blob][r]{\cite{ronneberger2015unet}} & 7.9 & 83.3 & 63.89$\pm$1.30 & 
\makebox[\nameblob][l]{U-Net$_{15}$}
\makebox[\blob][r]{\cite{ronneberger2015unet}} & 16.3 & 800.9 & 74.47$\pm$1.57  
\\
\makebox[\nameblob][l]{PraNet$_{20}^{*}$} \makebox[\blob][r]{\cite{fan2020pranet}} & 30.5 & 15.7 & 82.56$\pm$1.08 &
\makebox[\nameblob][l]{UNETR$_{21}^{*}$} \makebox[\blob][r]{\cite{hatamizadeh2022unetr}} & 92.6 & 72.6 & 81.14$\pm$0.85
\\
\makebox[\nameblob][l]{Res2UNet$_{21}$} \makebox[\blob][r]{\cite{gao2019res2net}} & 25.4 & 17.8 & 81.62$\pm$0.97 &
\makebox[\nameblob][l]{Res2UNet$_{21}$} \makebox[\blob][r]{\cite{gao2019res2net}} & 38.3 & \textbf{44.2} & 79.23$\pm$0.66 \\
\hline
\SetCell[c=8]{l} \textit{Implicit Approaches} \\
\hline 
\makebox[\nameblob][l]{OSSNet$_{21}$} \makebox[\blob][r]{\cite{Reich2021OSSNetME}} & 5.2 & 6.4 & 76.11$\pm$1.14 & 
\makebox[\nameblob][l]{OSSNet$_{21}$} \makebox[\blob][r]{\cite{Reich2021OSSNetME}} & 7.6 & 55.1 & 73.38$\pm$1.65 \\
\makebox[\nameblob][l]{IOSNet$_{22}$} \makebox[\blob][r]{\cite{Khan2022IOSNet}} & 4.1 & \textbf{5.9} & 78.37$\pm$0.76 & 
\makebox[\nameblob][l]{IOSNet$_{22}$} \makebox[\blob][r]{\cite{Khan2022IOSNet}} & 6.2 & 46.2 & 76.75$\pm$1.37 \\
\hline 
SwIPE (\textit{ours}) & \textbf{2.7} & 10.2 & \textbf{85.05}$\pm$0.82 &
SwIPE (\textit{ours}) & \textbf{4.4} & 71.6 & \textbf{81.21}$\pm$0.94 \\
% \cline{4-5}
\end{tblr}
}
% \vspace*{-6mm}
\end{center}
\vspace*{-10mm}
\end{table*}

\vspace{-3mm}
\subsection{Training SwIPE} \label{sec:2-3}
\vspace{-1.5mm}

To optimize the parameters of $f_\theta$, we first sample a set of point and occupancy pairs $\{\textbf{p}^\mathbb{S}_i, \textbf{o}_i\}_{i \in \mathcal{I}}$ for each source image, where $\mathcal{I}$ is the index set for the selected points.
We obtain an equal number of points for each foreground class using Latin Hypercube sampling with 50\% of each class's points sampled within 10 pixels of the class object boundaries. 
The \textbf{point-wise occupancy loss}, written as 
$\mathcal{L}_\text{occ}(\textbf{o}_i, \hat{\textbf{o}}_i) = 0.5 \cdot \mathcal{L}_\text{ce}(\textbf{o}_i, \hat{\textbf{o}}_i) + 0.5 \cdot \mathcal{L}_\text{dc}(\textbf{o}_i, \hat{\textbf{o}}_i)$,
is an equally weighted sum of Cross Entropy loss $\mathcal{L}_\text{ce}(\textbf{o}_i, \hat{\textbf{o}}_i) = -\log \hat{o}_i^c$ and Dice loss $\mathcal{L}_\text{dc}(\textbf{o}_i, \hat{\textbf{o}}_i) = 1 - \frac{1}{C} \sum_{c} \frac{2 \cdot o_i^c \cdot \hat{o}_i^c + 1}{(o_i^c)^2 + (\hat{o}_i^c)^2 + 1}$, where $\hat{o}_i^c$ is the predicted probability for the target occupancy with class label $c$.
Note that in practice, these losses are computed in their vectorized forms; for example, the Dice loss is applied with multiple points per image instead of an individual point (similar to computing the Dice loss between a flattened image and its flattened mask).
% The class occupancy loss for a point consists of an equally weighted sum of Cross Entropy loss $\mathcal{L}_\text{ce}$ and Dice Loss $\mathcal{L}_\text{dc}$ and can be written as  
% where }_i^c$ is the predicted probability for the target occupancy target with class label $c$.
% The class occupancy loss for a point consists of an equally weighted sum of Cross Entropy loss $\mathcal{L}_\text{ce}$ and Dice Loss $\mathcal{L}_\text{dc}$ and can be written as  
% $\mathcal{L}_\text{occ}(\textbf{o}_i, \hat{\textbf{o}}_i) = 0.5 \cdot \mathcal{L}_\text{ce}(\textbf{o}_i, \hat{\textbf{o}}_i) + 0.5 \cdot \mathcal{L}_\text{dc}(\textbf{o}_i, \hat{\textbf{o}}_i)$, 
% where 
% $\mathcal{L}_\text{ce}(\textbf{o}_i, \hat{\textbf{o}}_i) = -log \hat{o}_i^c$,  
% $\mathcal{L}_\text{dc}(\textbf{o}_i, \hat{\textbf{o}}_i) = 1 - \frac{1}{C} \sum_{c} \frac{2 \cdot o_i^c \cdot \hat{o}_i^c + 1}{(o_i^c)^2 + (\hat{o}_i^c)^2 + 1}$, 
% and $\hat{o}_i^c$ is the predicted probability for the target occupancy target with class label $c$.
The \textbf{loss for patch and image decoder} predictions is $\mathcal{L}_{\mathbb{P}\mathbb{I}}(\textbf{o}_i, \hat{\textbf{o}}_i^\mathbb{P}, \hat{\textbf{o}}_i^\mathbb{I}) = \alpha \mathcal{L}_\text{occ}(\textbf{o}_i, \hat{\textbf{o}}_i^\mathbb{P}) + (1 - \alpha) \mathcal{L}_\text{occ}(\textbf{o}_i, \hat{\textbf{o}}_i^\mathbb{I})$, where $\alpha$ is a local-global balancing coefficient.
Similarly, the \textbf{loss for the SPO} occupancy prediction $\hat{\textbf{o}}_i'$ is $\mathcal{L}_{\text{SPO}}(\textbf{o}_i, \hat{\textbf{o}}_i') = \mathcal{L}_\text{occ}(\textbf{o}_i, \hat{\textbf{o}}_i')$. 
Finally, the \textbf{overall loss} for a coordinate is formulated as 
$\mathcal{L} = \mathcal{L}_{\mathbb{P}\mathbb{I}} + \beta \mathcal{L}_{\text{SPO}} + \lambda(||\textbf{z}_i^\mathbb{P}||_2^2 + ||\textbf{z}_i^\mathbb{I}||_2^2)$, where $\beta$ scales SPO and the last term (scaled by $\lambda$) regularizes the patch \& image embeddings, respectively.
% $\mathcal{L} = \mathcal{L}_{\mathbb{P}\mathbb{I}} + \beta \mathcal{L}_{\text{SPO}} + \lambda(||\textbf{z}_i^\mathbb{P}||_2^2 + ||\textbf{z}_i^\mathbb{I}||_2^2 + ||\textbf{z}_i^`||_2^2)$, where $\beta$ scales SPO and the last term regularizes the patch, image, and SPO patch embeddings, respectively.

\vspace{-3mm}
\section{Experiments and Results} \label{sec:3}
\vspace{-1mm}

This section presents quantitative results from \textbf{four main studies}, analyzing overall performance, robustness to data shifts, model \& data efficiency, and ablation \& component studies.
For more implementation details, experimental settings, and qualitative results, we refer readers to the Supplementary Material. 

% This section presents quantitative and qualitative results with \textbf{four main studies}.
% 1) Overall performance compared to state-of-the-art implicit and discrete methods.
% 2) Robustness to data shifts across different image resolutions, datasets, and modalities. 
% 3) Model and data efficiency with varying backbone sizes and annotation availability.
% 4) Ablations and component choices to study the contributions and design choices for each proposed component.

\vspace{-4mm}
\subsection{Datasets, Implementations, and Baselines} \label{sec:3-1}
\vspace{-1.5mm}

We evaluate performance on two tasks: 2D binary polyp segmentation and 3D multi-class abdominal organ segmentation. 
For polyp segmentation, we train on the challenging \textbf{Kvasir-Sessile} dataset \cite{jha2021sessilecomprehensive} (196 colored images of small sessile polyps), and use \textbf{CVC-ClinicDB} \cite{bernal2015cvc} to test model robustness.  
For 3D organ segmentation, we train on \textbf{BCV} \cite{bcv2015} (30 CT scans, 13 annotated organs), and
use the diverse CT images in \textbf{AMOS} \cite{ji2022amos} (200 training CTs, the same setting used in~\cite{zhang2022spade}) to evaluate model robustness. 
% given its diverse acquisition profile.
All the datasets are split with a 60:20:20 train:validation:test ratio. 
For each image in Sessile [in BCV, resp.], we obtain 4000 [20,000] background points and sample 2000 [4000] foreground points for each class with half of every class' foreground points lying within 10 pixels [voxels] of the boundary.

2D Sessile Polyp training uses a modified Res2Net \cite{gao2019res2net} backbone with 28 layers, [256, 256, 256] latent MLP dimensions for $D^\mathbb{P}$, [256, 128] latent dimensions for $D^\mathbb{I}$, $d=128$, $S=32$, and $con=8$.
3D BCV training uses a Res2Net-50 backbone, [256, 256, 256, 256] latent MLP dimensions for $D^\mathbb{P}$, [256, 256, 128] latent MLP dimensions for $D^\mathbb{I}$, $d=512$, $S=8$, and $con=6$ (all adjoining patches in 3D).
The losses for both tasks are optimized with AdamW~\cite{Loshchilov2017DecoupledWDAdamW} and use $\alpha$=$0.5$, $\beta$=$0.1$, and $\lambda$=$0.0001$.
For inference, we adopt MISE like prior works~\cite{Mescheder2018OccNet,Khan2022IOSNet,Reich2021OSSNetME} and evaluate on a reconstructed prediction mask equal in size to the input image.
$D^\mathbb{P}$ segments boundaries better than $D^\mathbb{I}$, and is used to produce final predictions. 
% We refer readers to the supplementary section for more details.

For fair comparisons, all the methods are trained using the same equally-weighted Dice and Cross Entropy loss for 30,000 and 50,000 iterations on 2D Sessile and 3D BCV, resp.
The test score at the best validation epoch is reported. 
Image input sizes were $384\times384$ for Sessile and $96\times96\times96$ for BCV.
All the implicit methods utilize the same pre-sampled points for each image.
For IOSNet~\cite{Khan2022IOSNet}, both 2D and 3D backbones were upgraded from three downsampling stages to five for fair comparisons and empirically confirmed to outperform the original. 
We omit comparisons against IFA~\cite{hu2022ifanet} to focus on medical imaging approaches; plus, IFA did not outperform IOSNet~\cite{Khan2022IOSNet} on either task.

% EXPERIMENTS
\begin{table}[!tb]
    \caption{ \label{tab:2}
    \textbf{Left} and \textbf{Middle}: Robustness to data shifts. 
    \textbf{Right}: Efficiency studies.}
    \vspace*{-3mm}
    \begin{minipage}{.31\linewidth}
        \hspace{5mm}\text{Across Resolutions}\vspace{0.5mm}
      \resizebox{0.98\textwidth}{!}{
        \begin{tblr}{
        columns={colsep=3pt},
        colspec={c l | c c },
        row{1} = {gray!50!black!15},
        row{2,9} = {gray!20!black!5}
        }
            \hline
            & Method & Size & Dice \\
            \hline
            \SetCell[c=4]{l} \textit{Varying Output Size} \\
            1 & PraNet \cite{fan2020pranet}  & 128$\downarrow$ & 72.64 \\
            2 & IOSNet \cite{Khan2022IOSNet} & 128$\downarrow$ & 76.18 \\
            3 & SwIPE & 128$\downarrow$ & 81.26 \\
            \hline 
            4 & PraNet \cite{fan2020pranet} & 896$\uparrow$ & 74.95 \\
            5 & IOSNet \cite{Khan2022IOSNet} & 896$\uparrow$ & 78.01 \\
            6 & SwIPE & 896$\uparrow$ & 84.33 \\
            \hline
            \SetCell[c=4]{l} \textit{Varying Input Size} \\
            7 & PraNet \cite{fan2020pranet}  & 128$\downarrow$ & 68.79  \\
            8 & PraNet \cite{fan2020pranet}  & 896$\uparrow$  & 43.92  \\
            \end{tblr}
    }
    \end{minipage}
    \begin{minipage}{.31\linewidth}
            \centering
        \hspace{0.8mm}\text{Across Datasets}\vspace{0.5mm}
        \resizebox{0.85\textwidth}{!}{               
            \begin{tblr}{
        columns={colsep=2pt},
        colspec={c l | c},
        row{1} = {gray!50!black!15},
        row{2,6,7} = {gray!20!black!5}
        }
        \hline
        & Method & Dice \\
        \hline
        \SetCell[c=3]{l} \textit{Polyp Sessile \hspace{-1.2mm} $\rightarrow$ \hspace{-1.6mm} CVC} \\
        1 & PraNet~\cite{fan2020pranet}  & 68.37 \\
        2 & IOSNet~\cite{Khan2022IOSNet} & 59.42 \\
        3 & SwIPE & 70.10 \\
        \hline
        \SetCell[c=3]{l} {\small \textit{CT BCV \hspace{-1.2mm} $\rightarrow$ \hspace{-1.6mm} CT AMOS}} \\   [-4pt]
        \SetCell[c=3]{l} \textit{(liver class only)} \\ 
        4 & UNETR~\cite{hatamizadeh2022unetr} & 81.75 \\
        5 & IOSNet~\cite{Khan2022IOSNet} & 79.48 \\
        6 & SwIPE & 82.81 \\
        \end{tblr}
        }
    \end{minipage} 
    \begin{minipage}{.31\linewidth}
        % \hspace{7mm}
        % \text{Model and Data Efficiency}
        % \resizebox{0.8\textwidth}{!}{               
            \includegraphics[width=\textwidth]{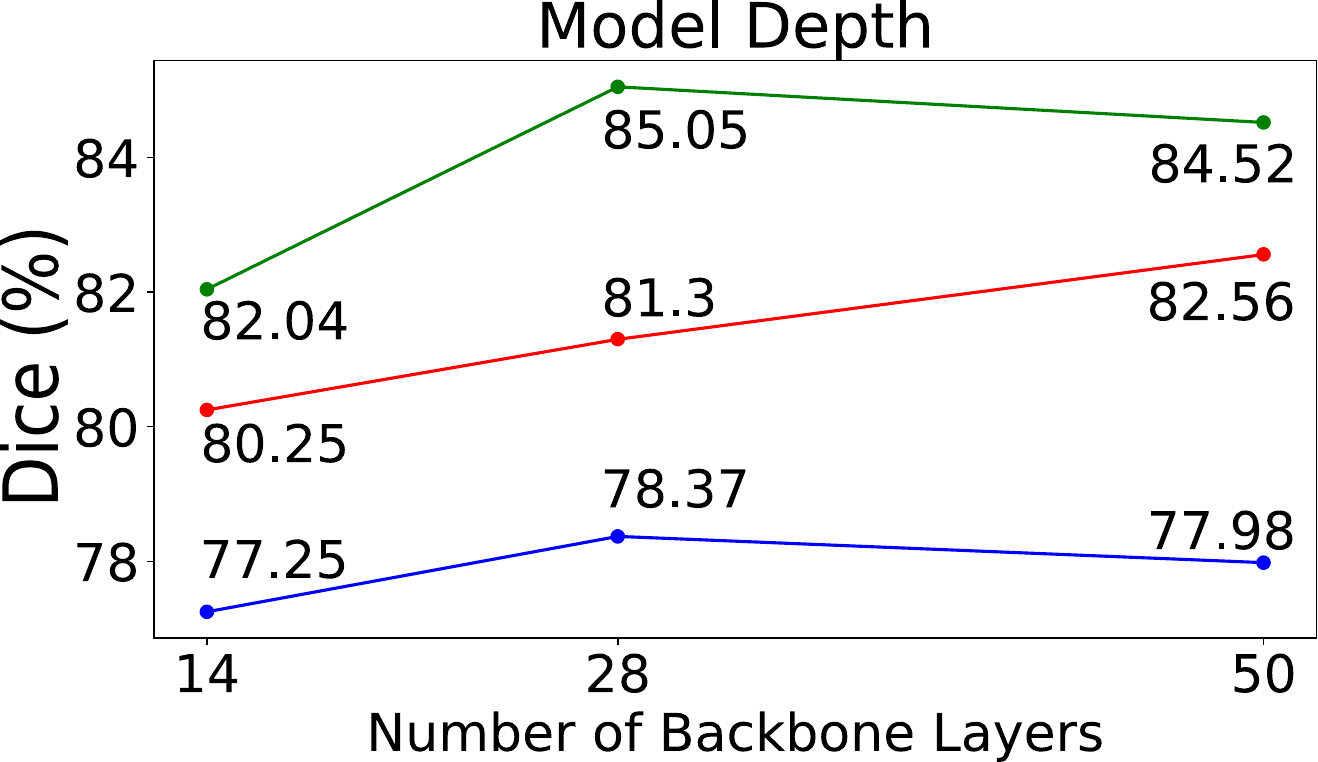} 
            \includegraphics[width=\textwidth]{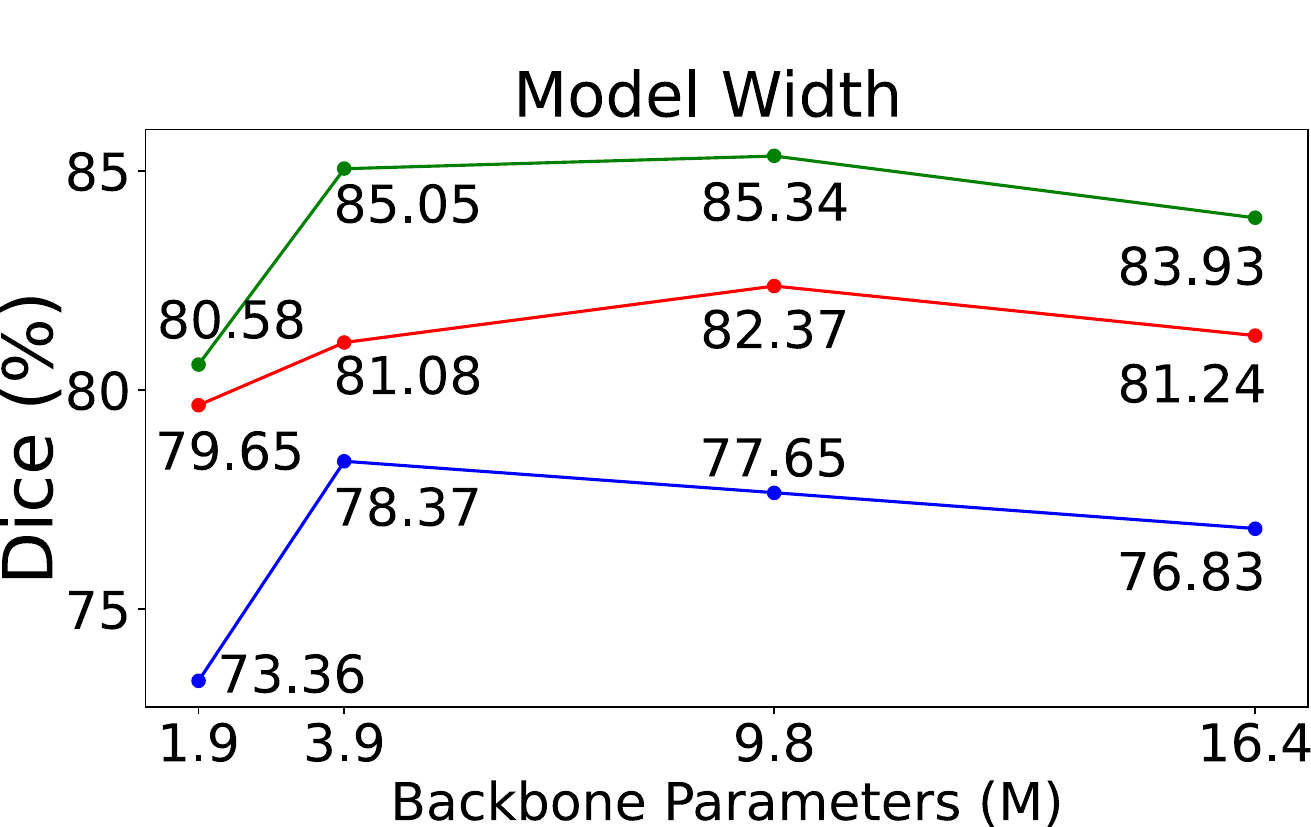} 
            \includegraphics[width=\textwidth]{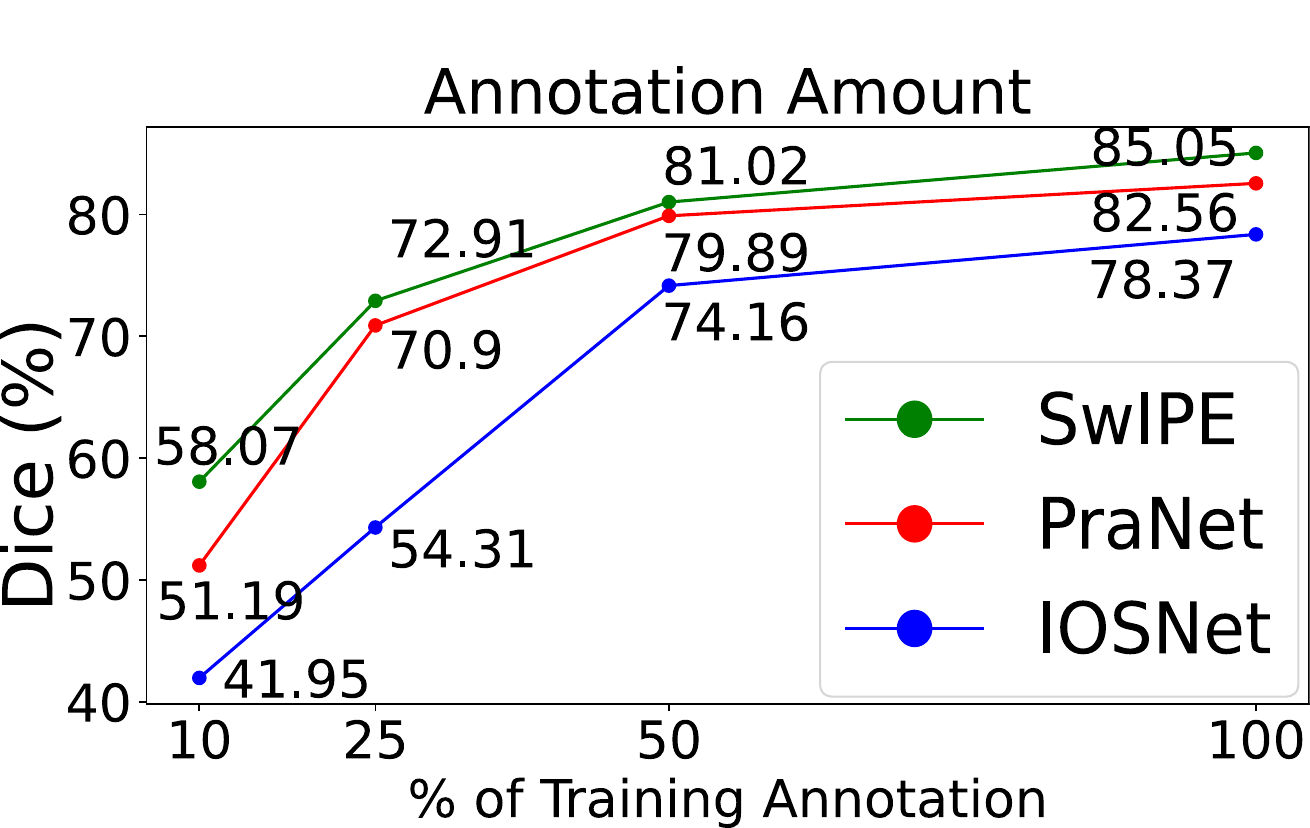}

        % }
    \end{minipage} 
    \vspace*{-7mm}
\end{table}

\vspace{-4mm}
\subsection{Study 1: Performance Comparisons} \label{sec:3-2}
\vspace{-1.5mm}

The results for 2D Polyp Sessile and 3D CT BCV organ segmentation are presented in Table~\ref{tab:1}.
FLOPs are reported from the forward pass on a single image during training. 

On the smaller polyp dataset, we observe notable improvements over the best-known implicit approaches (+6.7\% Dice) and discrete methods (+2.5\% Dice) with much fewer parameters (9\% of PraNet~\cite{fan2020pranet} and 66\% of IOSNet~\cite{Khan2022IOSNet}).
For BCV, the performance gains are more muted; however, we still marginally outperform UNETR~\cite{hatamizadeh2022unetr} with over 20x fewer parameters and comparable FLOPs.

\vspace*{-5mm}
\subsection{Study 2: Robustness to Data Shifts} \label{sec:3-3}
\vspace*{-1.5mm}

% We explore the robustness of different methods to image resolution changes and dataset distribution shifts (see left and middle of Table~\ref{tab:2})
In this study, we explore the robustness of various methods to specified target resolutions and dataset shifts.
The left-most table in Table~\ref{tab:2} contains results for the former study conducted on 2D Sessile, where we first analyze the effect of directly resizing outputs (Tab.~\ref{tab:2} left, rows 1 to 6) when given an input image that is standard during training ($384\times384$). 
The discrete method, PraNet, outputs $384\times384$ predictions which are interpolated to the target size (Tab.~\ref{tab:2} left, rows 1 \& 4).
This causes more performance drop-offs than the implicit methods which can naturally vary the output size by changing the resolution of the input coordinates.
We also vary the input size so that no manipulations of predictions are required (Tab.~\ref{tab:2} left, rows 7 \& 8), which results in steep accuracy drops.

The results for the dataset shift study are given in the middle of Table~\ref{tab:2}, where CVC is another binary poly segmentation task and the liver class is evaluated on all CT scans in AMOS.
Both discrete methods outperform IOSNet, which may indicate that point-based features are more prone to overfitting due to a lack of contextual regularization. 
Also, we highlight our method's consistent outperformance over both discrete methods and IOSNet in all of the settings.

% , we study the robustness of the best discrete and implicit method on the Polyp Sessile task to varying prediction resolutions.
% First, we analyze the effect of resizing outputs (rows 1 to 6) and input the standard $384\times384$ image size seen during training.
% For PraNet, $384\times384$ predictions are outputted and then resized to the desired output size, while implicit methods vary output size by changing the number of predicted coordinates.
% Implicit methods demonstrate their spacial flexibility and maintain high performances.
% Between implicit approaches, our patch-based schema yields better performance across the board.
% Second, we analyze the effect of varying the input size to match the target output size.
% We omit implicit methods here since they can predict masks at arbitrary resolutions using their training resolution.
% We observe steep performance declines with the discrete method probably due to drastic content differences in usual receptive fields.

% Further we study robustness to data shifts in the same task across datasets.
% For polyp segmentation, we see general outperformance of PraNet over IOSNet.
% Limitations of point-wise feature aggregation over those that aggregate features specially (i.e. convolutions).
% For 3D segmentation, we predict the liver class in AMOS CT images to gauge shape realism. 

\begin{table}[!tb]
    \caption{ \label{tab:3}
    \textbf{Left}: Ablation studies.
    \textbf{Right}: Design choice experiments.
    \vspace*{-10mm}
    }
    \vspace{-2mm}
    \begin{minipage}{.65\linewidth}
      \hspace{2mm}\text{Ablation Studies on 2D Sessile}\vspace{0.75mm}
        \centering
      \resizebox{0.95\textwidth}{!}{
        \begin{tblr}{
        columns={colsep=3pt},
        colspec={c c | c c c c c c c c},
        row{1} = {gray!50!black!15},
        row{9} = {gray!20!black!5}
        }
            \SetCell[c=2]{c} Component & & \SetCell[c=8]{c} Incorporation &&&&&&& \\
            \hline 
            % \SetCell[c=4]{l} \textit{Varying Resolutions} \\
            $E_n$ & RFB-Lite        & & \cmark & \cmark & \cmark & \cmark & \cmark & \cmark & \cmark \\
            $E_n$ & Cascade         & &        & \cmark & \cmark & \cmark & \cmark & \cmark & \cmark \\
            $E_n$ & MEA             & &        &        & \cmark & \cmark & \cmark & \cmark & \cmark \\
            \hline 
            $D^\mathbb{I}$ & $\textbf{z}^\mathbb{I}, \textbf{p}^\mathbb{I}$ & 
                                    &        &        &  & \cmark & \cmark & \cmark & \cmark \\
            $D^\mathbb{P}$ & $\textbf{z}^\mathbb{I}, \textbf{p}^\mathbb{I}$ & 
                                    &        &        &        &   & \cmark & \cmark & \cmark \\
            $D^\mathbb{P}$ & $\textbf{p}^\mathbb{S}$ & 
                                    &        &        &        &        &   & \cmark & \cmark \\
            $D^\mathbb{P}$ & SPO &  &        &        &        &        &        &  & \cmark \\
            \hline 
            \SetCell[c=2]{c} Dice (\%) & & 76.44 & 76.57 & 78.19 & 80.33 & 80.92 & 82.28 & 83.75 & \textbf{85.05}\\
            \end{tblr}
    }
    \end{minipage}%
    \begin{minipage}{.35\linewidth}
        \hspace{2mm}\text{Alternative Designs}\vspace{0.5mm}
        \centering
        \resizebox{0.85\textwidth}{!}{               
            \begin{tblr}{
        columns={colsep=2.6pt},
        colspec={c l | c},
        row{1} = {gray!50!black!15},
        row{2,5,9} = {gray!20!black!5}
        }
            \SetCell[c=2]{c} Description & & Dice (\%) \\
            \hline
            % 1 & SwIPE & 85.05 \\
            % \hline
            \SetCell[c=3]{l} \textit{Backbone} \\
            1 & CCT~\cite{hassani2021CCT} & 78.30 \\
            2 & U-Net~\cite{ronneberger2015unet} & 79.94 \\
            \hline
            \SetCell[c=3]{l} \textit{MEA Replacements for Feature Fusion} \\
            3 & Addition         & 84.19 \\
            4 & Concat. + 1x1 Conv & 83.58 \\
            5 & Self-Attention   & 68.23 \\
            \hline
            \SetCell[c=3]{l} \textit{SPO} \\
            6 & $N_o=0$          & 83.75  \\
            7 & $N_o=4, Con=4$   & 83.71 \\
            8 & $N_o=4, Con=8$   & 83.94 \\
            9 & $N_o=8, Con=4$   & 84.43 \\
            \end{tblr}
            }
    \end{minipage} 
    \vspace*{-6.5mm}
\end{table}

\vspace*{-4mm}
\subsection{Study 3: Model Efficiency and Data Efficiency} \label{sec:3-4}
\vspace*{-1.5mm}

% In medical image segmentation, model efficiency is important due to the limited computation budgets of many clinics or labs while data efficiency is paramount given the scarcity of annotation in medical data. 
To analyze the model efficiency (the right-most column of charts in Table~\ref{tab:2}), we report on 2D Sessile and vary the backbone size in terms of depth and width.
For data efficiency, we train using 10\%, 25\%, 50\%, and 100\% of annotations. 
% Experiments were conducted on 2D polyp sessile segmentation which often involves an examination using devices with limited compute and where labels are expensive to obtain.
Not only do we observe outperformance across the board in model sizes \& annotation amounts, but the performance drop-off is more tapered with our method.

\vspace*{-4mm}
\subsection{Study 4: Component Studies and Ablations} \label{sec:3-5}
\vspace*{-1.5mm}

%Finally, 
The left side of Table~\ref{tab:3} presents our ablation studies, showing the benefits enabled by context aggregation within $E_n$, global information conditioning, and adoption of MEA \& SPO. 
We also explore alternative designs on the right side of Table~\ref{tab:3} for our three 
%most important 
key components, and affirm their contributions 
%toward 
on achieving superior performance.

\vspace*{-2.5mm}
\section{Conclusions} \label{sec:4}
\vspace*{-1.25mm}

% To address this gap, we present SwIPE, a novel approach that utilizes implicit patch embeddings to achieve efficient and robust medical image segmentation. SwIPE directly models object shapes and adopts continuous representations, enabling accurate segmentation while avoiding the limitations of traditional approaches.

SwIPE represents a notable departure from conventional discrete segmentation approaches and directly models object shapes instead of pixels and utilizes continuous rather than discrete representations. 
By adopting both patch and image embeddings, our approach enables accurate local geometric descriptions and improved shape coherence.
% Our approach also adopts both patch and image embeddings to enable the efficient representation of local object geometry and appearance.
% Our approach is based on the use of implicit patch embeddings that enable the representation of local object geometry and appearance in an expressive and memory-efficient manner. 
% SwIPE is designed to work with 
% %a variety of 
% different medical image modalities and has been
% %extensively 
% well evaluated on several benchmark datasets. 
Experimental results show the superiority of SwIPE over state-of-the-art approaches in terms of segmentation accuracy, efficiency, and robustness. 
The use of local INRs represents a new direction for medical image segmentation, and we hope to inspire further research in this direction.

\vspace*{-1mm}

\bibliographystyle{splncs04}
\bibliography{refs}

%%%%%%% ------------------------------------------------------------------------- %%%%%%%
%%%%%%% ------------------------------------------------------------------------- %%%%%%%

\end{document}